\documentclass[10pt,twocolumn,letterpaper]{article}

\usepackage[pagenumbers]{cvpr} 
\usepackage{lipsum}
\usepackage{float}
\definecolor{cvprblue}{rgb}{0.21,0.49,0.74}
\usepackage[pagebackref,breaklinks,colorlinks,allcolors=cvprblue]{hyperref}
\usepackage{multirow}
\usepackage{graphicx}

\usepackage{amsmath} 
\usepackage{algorithm}
\usepackage{algpseudocode}
\usepackage{amsfonts} 

\def\paperID{7093}
\def\confName{ICCV}
\def\confYear{2025}
\def\paperTitle{DetailGen3D: Generative 3D Geometry Enhancement via Data-Dependent Flow}

\def\authorBlock{
    Ken Deng\textsuperscript{1,2,3} \quad
    Yuan-Chen Guo\textsuperscript{2} \quad
    Jingxiang Sun\textsuperscript{1} \quad
    Zi-Xin Zou\textsuperscript{2} \quad
    Yangguang Li\textsuperscript{2} \quad
    Xin Cai\textsuperscript{4} \\[0.5em]
    Yan-Pei Cao\textsuperscript{2} \quad
    Yebin Liu\textsuperscript{1} \quad
    Ding Liang\textsuperscript{1,2} \\[0.5em]

    \small
    \textsuperscript{1}Tsinghua University \quad
    \textsuperscript{2}VAST \quad
    \textsuperscript{3}Sun Yat-sen University
    \textsuperscript{4}The Chinese University of Hong Kong
}
\definecolor{darkred}{rgb}{0.7, 0.0, 0.0}
\definecolor{darkred2}{rgb}{0.5, 0.0, 0.0}
\definecolor{darkred3}{rgb}{0.9, 0.0, 0.0}
\definecolor{darkgreen}{rgb}{0.0, 0.42, 0.24}
\definecolor{darkblue}{rgb}{0.10, 0.17, 0.8}
\definecolor{Gray}{gray}{0.93}

\definecolor{coarse_color}{RGB}{140, 146, 167}
\definecolor{fine_color}{RGB}{153, 77, 77}

\begin{document}

\title{\paperTitle}
\author{\authorBlock}

\twocolumn[{
\renewcommand\twocolumn[1][]{#1}
\maketitle
\begin{center}
    \captionsetup{type=figure}
    \includegraphics[width=\textwidth]{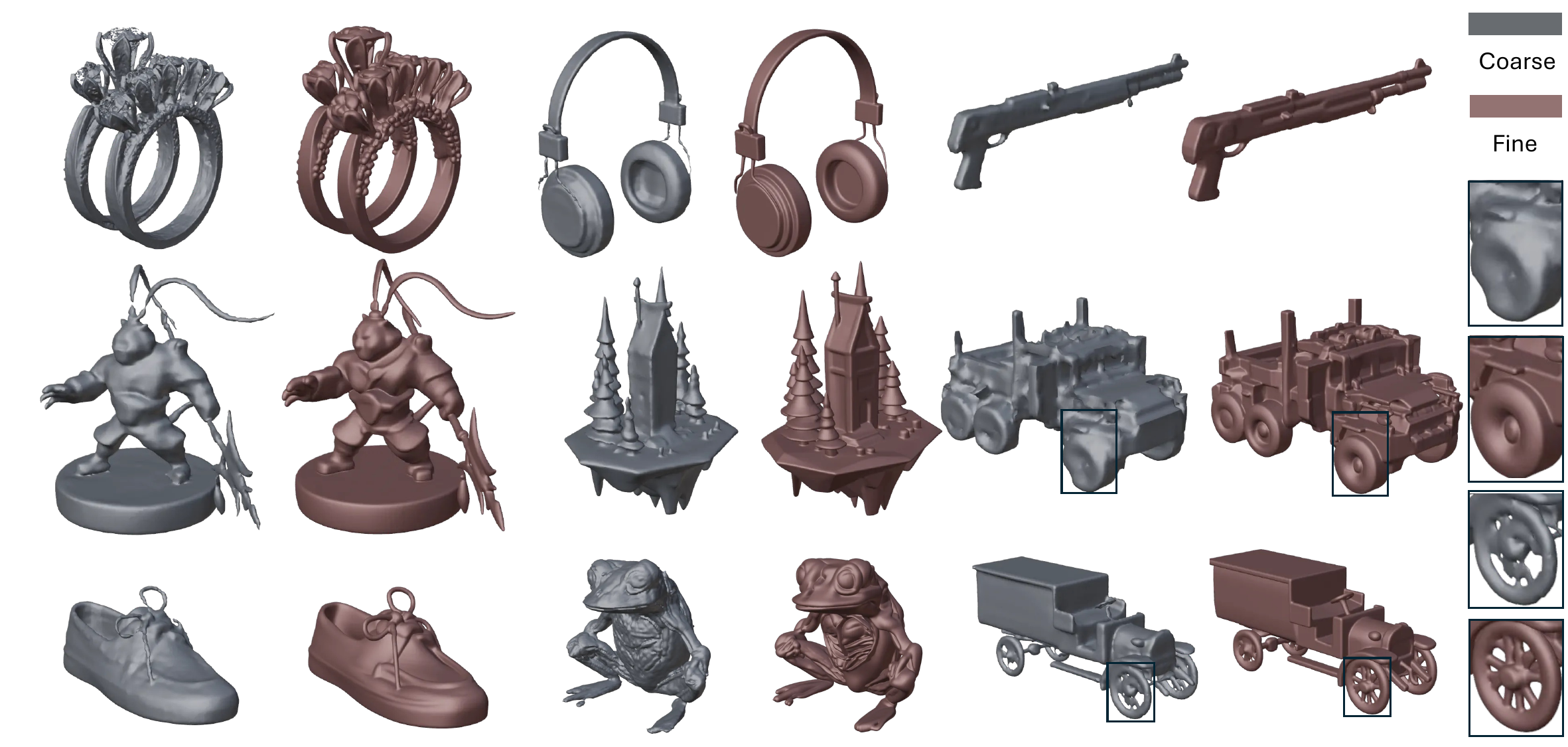} 
    \captionof{figure}{Our method demonstrates effective geometry refinement across various tasks and representations. In the images, the coarse geometry is displayed in gray \textcolor{coarse_color}{\rule{0.8em}{0.8em}}, while the refined geometry produced by our approach is shown in red \textcolor{fine_color}{\rule{0.8em}{0.8em}}. On the right, zoomed-in details are provided to better observe the refinement effects.}
    \label{fig:teaser}
\end{center}
}]


\begin{abstract}
Modern 3D generation methods can rapidly create shapes from sparse or single views, but their outputs often lack geometric detail due to computational constraints.  We present DetailGen3D, a generative approach specifically designed to enhance these generated 3D shapes. Our key insight is to model the coarse-to-fine transformation directly through data-dependent flows in latent space, avoiding the computational overhead of large-scale 3D generative models. We introduce a token matching strategy that ensures accurate spatial correspondence during refinement, enabling local detail synthesis while preserving global structure. By carefully designing our training data to match the characteristics of synthesized coarse shapes, our method can effectively enhance shapes produced by various 3D generation and reconstruction approaches, from single-view to sparse multi-view inputs. Extensive experiments demonstrate that DetailGen3D achieves high-fidelity geometric detail synthesis while maintaining efficiency in training. Our project page is \url{https://detailgen3d.github.io/DetailGen3D/}
\end{abstract}

\vspace{-2em}
\section{Introduction}
\label{sec:intro}

\setlength{\parindent}{2em}
Obtaining high-quality 3D geometry has been a long-standing research focus in the fields of computer vision and graphics. High-quality 3D models are not only valuable in the film industry, video games, and virtual reality but also play a crucial role in the rapidly advancing field of embodied intelligence, contributing significantly to simulation environments. Early approaches for high-quality multi-view stereo reconstruction rely on dense multi-view inputs~\cite{goesele2007multi,furukawa2009accurate,agarwal2011building, yao2018mvsnet,huang2018deepmvs,chen2019point}. Although recent methods~\cite{wang2023neuslearningneuralimplicit,wang2023neus2fastlearningneural,Huang2DGS2024,chen2024pgsrplanarbasedgaussiansplatting} based on Neural Radiance Field (NeRF)~\cite{mildenhall2021nerf} and 3D Gaussian Splatting (3DGS)~\cite{kerbl3Dgaussians} have improved the end-to-end performance, reconstructing high-quality geometry remains a challenging problem.

Recent methods for 3D generation from sparse or single views have evolved into two main paradigms: 1) optimization using 2D generative models and 2) direct training with 3D data. The first approach leverages pre-trained 2D vision models - notably, DreamFusion ~\cite{poole2022dreamfusiontextto3dusing2d} introduced Score Distillation Sampling (SDS) to align rendered views with text-conditioned distributions. Related methods ~\cite{wang2023prolificdreamerhighfidelitydiversetextto3d,sun2023dreamcraft3d} have extended this paradigm, though they consistently face geometric inconsistency issues, such as the ``Janus problem'' where details conflict across viewpoints.
The second paradigm, training with 3D data, has seen multiple technical advances. Multi-view diffusion approaches ~\cite{liu2023zero1to3,liu2023one2345singleimage3d,liu2023one2345fastsingleimage,shi2024mvdreammultiviewdiffusion3d,li2023sweetdreameraligninggeometricpriors,qiu2023richdreamergeneralizablenormaldepthdiffusion,liu2024syncdreamergeneratingmultiviewconsistentimages,long2023wonder3dsingleimage3d,shi2023zero123singleimageconsistent,wang2023imagedreamimagepromptmultiviewdiffusion} achieve view consistency but struggle with geometric coherence in fine details. Feed-forward Large Reconstruction Models ~\cite{hong2024lrmlargereconstructionmodel,xu2023dmv3ddenoisingmultiviewdiffusion,li2023instant3dfasttextto3dsparseview,wang2023pflrmposefreelargereconstruction,zou2024triplane,liu2024meshformer,wei2024meshlrmlargereconstructionmodel,zhang2024gslrmlargereconstructionmodel} offer impressive speed but face resolution constraints. While recent 3D diffusion models ~\cite{zhang20233dshape2vecset3dshaperepresentation,zhao2023michelangeloconditional3dshape,zhang2024claycontrollablelargescalegenerative,wu2024direct3d,ren2024xcubelargescale3dgenerative,zhang2024gaussiancubestructuredexplicitradiance} demonstrate quality improvements through direct 3D training, computational demands continue to limit their achievable resolution and detail.

While recent advances in 3D generation have shown promising results, the generated shapes often lack fine geometric details. Simply scaling up these generative models demands prohibitive computational resources. Traditional geometry refinement approaches that rely on high-resolution dense multi-view images are ill-suited for enhancing such generated shapes, as these additional inputs are typically unavailable. Furthermore, existing optimization-based methods that utilize normal or shading information~\cite{wu2011high,yu2013shading,wu2011shading,xu2017shading,nehab2005efficiently} require precise texture alignment and struggle to preserve global shape coherence. To address these challenges, we propose \emph{DetailGen3D}, a generative approach to 3D geometry refinement that learns to synthesize plausible fine-scale details directly from high-quality 3D shapes. By capturing the underlying geometric patterns through data-driven priors, our method can enhance coarse shapes while maintaining structural consistency, even under noisy or incomplete information.

To tackle geometry refinement tasks with a generative model, a straightforward approach is to train a coarse geometry conditioned diffusion model, as seen in 2D image restoration methods \cite{cai2024phocolens, yu2024scalingexcellencepracticingmodel}. However, this method can be resource-intensive and may not be optimal for 3D tasks. Instead, inspired by Fischer et al. \cite{fischer2023boosting}, who achieved effective image super-resolution without relying on large generative models, we propose a more training-efficient strategy. Rather than learning a complex mapping from noise to fine details, we focus on directly modeling the transformation between coarse and fine geometry using data-dependent rectified flow \cite{liu2022flowstraightfastlearning}. This approach utilizes optimal transport, which provides a direct and structured mapping between coarse and fine geometry. By exploiting this coupling, we eliminate the need for the random coupling of noise and fine details, thus significantly reducing the training cost. This method is better suited for scaling up, offering a more efficient path to geometry refinement. 

Specifically, we introduce a training method called token matching. Establishing the refinement process locally is essential; otherwise, global refinement leads to inefficient training and hinders detail capturing. Token matching matches the coarse geometry latent code with the fine geometry latent code in latent space one-to-one, improving training efficiency. It prevents the network from learning unnecessary operations (e.g., swapping positions between two latent codes) and focuses solely on local refinement operations. This approach enables the capture of geometry details even with a small network.



Effective geometry refinement requires carefully designed training data. While high-quality 3D shapes are available, obtaining matching coarse-fine pairs poses a significant challenge. Simply applying traditional mesh degradation algorithms (e.g., simplification or smoothing operations) leads to simple objects remaining unchanged or extreme degradation on complex objects, resulting in low-quality coarse-fine pairs. To address this, we leverage an LRM-based model that reconstructs 3D geometry from sparse-view renderings of high-quality fine models. This approach enables consistent degradation across objects of varying complexity, enhances the utilization of existing 3D objects, and increases both the quality and quantity of coarse-fine pairs.


In summary, our contributions are as follows:
\begin{enumerate}
    \item We develop a novel generative geometry refinement algorithm, demonstrating its highly effectiveness for different geometric representations.
    
    \item We propose a data-dependent rectified flow to incorporate coarse geometry information, enabling local distribution transformation from coarse to refined geometry.
    
    \item We introduce a token matching training method that significantly enhances training efficiency and spatial correspondence accuracy.
    
\end{enumerate}

\section{Related Works}
\label{sec:related}

\setlength{\parindent}{2em}

\noindent\textbf{3D generation using 2D generative model.}
With the significant advancements in text-to-image generation models~\cite{rombach2021highresolution,saharia2022photorealistictexttoimagediffusionmodels,betker2023improving}, methods for text to 3D generation based on SDS loss optimization ~\cite{poole2022dreamfusiontextto3dusing2d,lin2023magic3dhighresolutiontextto3dcontent,xu2023dream3dzeroshottextto3dsynthesis,chen2023fantasia3ddisentanglinggeometryappearance,wang2023prolificdreamerhighfidelitydiversetextto3d,sun2023dreamcraft3d,tang2023dreamgaussian,yi2023gaussiandreamer,EnVision2023luciddreamer} have emerged, allowing the acquisition of 3D models without relying on multi-view inputs. However, these image diffusion models lack 3D priors, leading to the generation of 3D models that often suffer from the “Janus problem”, with poor alignment between geometry and texture, as well as substantial time overhead. 
\\
\noindent\textbf{3D generation using 3D data training.}
Finetuning image diffusion models using Objaverse~\cite{deitke2022objaverseuniverseannotated3d}, Objaverse-XL~\cite{deitke2023objaversexluniverse10m3d} to generate reasonably consistent multi-views with 3D consistency, which can then be input into multi-view reconstruction models to obtain 3D geometry ~\cite{liu2023zero1to3,liu2023one2345singleimage3d,liu2023one2345fastsingleimage,shi2024mvdreammultiviewdiffusion3d,li2023sweetdreameraligninggeometricpriors,qiu2023richdreamergeneralizablenormaldepthdiffusion,liu2024syncdreamergeneratingmultiviewconsistentimages,long2023wonder3dsingleimage3d,shi2023zero123singleimageconsistent,wang2023imagedreamimagepromptmultiviewdiffusion}. Some methods attempting to improve multi-view consistency~\cite{tang2024lgmlargemultiviewgaussian,xu2024grmlargegaussianreconstruction,wang2024crmsingleimage3d,xu2024instantmeshefficient3dmesh}, but geometry quality is even worse than previous methods since only 2D supervision is used. Using triplane as 3D representation, large-scale reconstruction model 
~\cite{hong2024lrmlargereconstructionmodel,li2023instant3dfasttextto3dsparseview,wang2023pflrmposefreelargereconstruction,xu2023dmv3ddenoisingmultiviewdiffusion,tochilkin2024triposrfast3dobject,sun2024dreamcraft3d,wei2024meshlrmlargereconstructionmodel,zou2024triplane,yang2024dreamcomposer} utilizing transformer architectures and triplane representations with fixed-view multi-view inputs, enabling reconstruction in just a few seconds. However, the extensive training time and GPU memory requirements of large reconstruction models severely limit their resolution. Generative models using triplane as 3D representation~\cite{gupta20233dgentriplanelatentdiffusion,zhang2024compress3dcompressedlatentspace,hong20243dtopialargetextto3dgeneration,shue20223dneuralfieldgeneration,wu2024direct3dscalableimageto3dgeneration} meet the same problem as well. Using point cloud as 3D representation improves the efficiency~\cite{jun2023shapegeneratingconditional3d,nichol2022pointegenerating3dpoint,zhang20233dshape2vecset3dshaperepresentation,zhao2023michelangeloconditional3dshape,zhang2024claycontrollablelargescalegenerative,zhao2025hunyuan3d20scalingdiffusion, li2025triposghighfidelity3dshape} but also brings high training costs. Some recent methods use voxel~\cite{ren2024xcubelargescale3dgenerative,liu2024meshformerhighqualitymeshgeneration,xiang2024structured} or gaussian~\cite{tang2024lgmlargemultiviewgaussian,zhang2024gslrmlargereconstructionmodel,zhang2024gaussiancubestructuredexplicitradiance} as 3D representations are also efficient.



\noindent\textbf{3D shape detailization.} SketchPatch~\cite{10.1145/3414685.3417816} uses a PatchGAN~\cite{Isola_2017_CVPR} discriminator to mimic the local style of a reference image, stylizing plain solid-lined sketches. DECORGAN~\cite{chen2021decorgan3dshapedetailization}, which employs PatchGAN to generate detailed voxel shapes from input coarse voxels, with the geometric style of the generated shape derived from a detailed reference voxel model. ShaDDR~\cite{chen2023shaddrinteractiveexamplebasedgeometry} enhances the generated geometry of DECORGAN by utilizing a 2-level hierarchical GAN and introducing texture generation. Meanwhile, DECOLLAGE~\cite{chen2024decollage} further improves shape detailization through structure-preserving losses and adaptive weighting of style and global discriminators. However, these methods primarily focus on enhancing the voxel resolution of low-quality inputs, which is different from our method: pushing the boundaries of detail enhancement beyond current state-of-the-art generative models, capturing finer details missed by coarser methods.


\section{Methods}
\label{sec:method}

\setlength{\parindent}{2em}

\begin{figure*}[htbp]
    \centering
    \captionsetup{type=figure}
    \includegraphics[width=\textwidth]{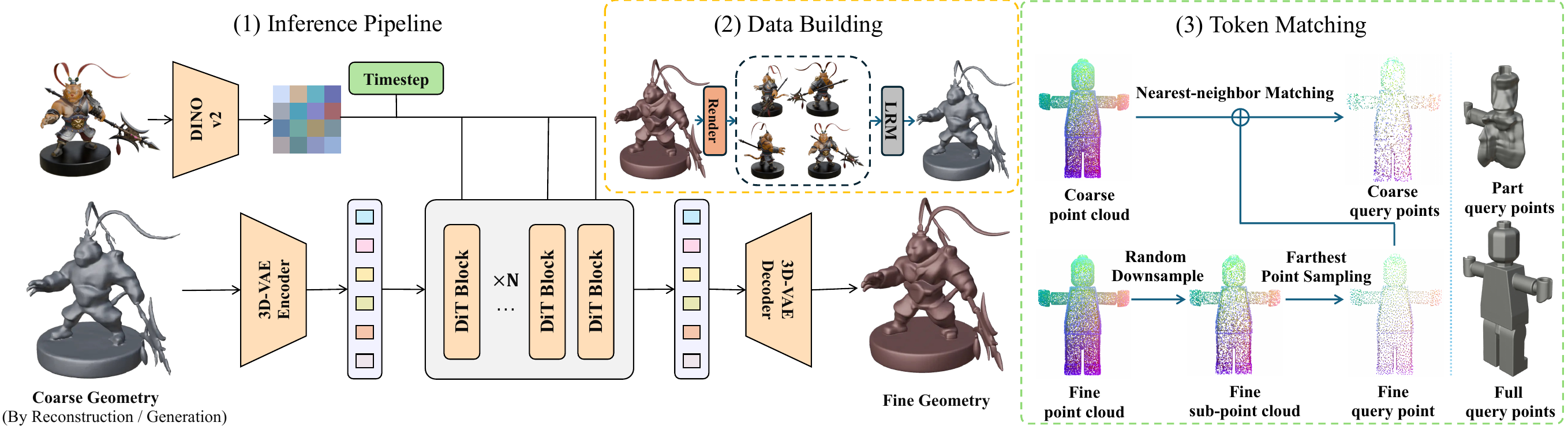} 
    \captionof{figure}{(1) Inference pipeline. We use 3D-VAE to extract tokens of the coarse geometry generated or reconstructed, then input the coarse token and DINOv2 feature~\cite{oquab2024dinov2learningrobustvisual} of the image prompt to DiT~\cite{Peebles2022DiT}. After the refinement process, we decode the predicted token using a 3D-VAE decoder to obtain refined geometry. The inference process takes only a few seconds. (2) For training data, we use reconstruction results reconstructed by LRM using multi-views rendered from fine geometry as coarse geometry. (3) We demonstrate the token matching process on the left. On the right, for the top one, we only use part query points, which are located in quadrant one, and for the bottom one, we use full query points, which demonstrate that tokens represent the space around the corresponding query points. }
    \label{fig:inference_pipeline}
    \vspace{-1.5em}
\end{figure*}

Our method \( F \) is designed to refine coarse geometry \( G_{\text{Coarse}} \) into fine geometry \( G_{\text{Fine}} \), guided by an image prompt \( I \). Here, \( G_{\text{Coarse}} \) may originate from reconstruction or generation processes, while \( G_{\text{Fine}} \) represents an enhanced version with improved surface quality and additional geometric details, such as noise removal and refinement:

\vspace{-5pt}
{\small \begin{equation}
    G_{Fine} = \mathcal{F}(G_{Coarse}, I)
\end{equation}}
\vspace{-5pt}

Considering this process has strong uncertainty, we model this process using a generative model. In particular, we select rectified flow~\cite{liu2022flowstraightfastlearning} because of its efficiency based on optimal transport and its ability to model the mapping relationship between two data distributions, i.e., between the distribution of coarse geometry and fine geometry. Our inference pipeline is shown in Fig~\ref{fig:inference_pipeline} (1).

\subsection{Data Dependent Rectified Flow}

We use data-dependent rectified flow to model the distribution mapping of coarse and fine geometry due to its efficiency. Compared to modeling the transformation between Gaussian noise and fine geometry, directly modeling the transformation between coarse and fine geometry can, to some extent, reconstruct the coarse shape without extensive denoising training, as demonstrated in Fig~\ref{fig:ablation_study}. In contrast, modeling the distribution mapping between noise and fine geometry, with coarse geometry as a condition via cross-attention, requires longer training time to achieve noise-to-coarse mapping.

\noindent\textbf{Model Architecture.} To improve efficiency, we build our refinement process in latent space, and our network consists of two parts: 3D Variational Autoencoder (3D-VAE) and Diffusion Transformer~\cite{Peebles2022DiT} (DiT).

\noindent\textbf{3D Variational Autoencoder.} The design of our 3D-VAE is primarily inspired by 3DShape2VecSet~\cite{zhang20233dshape2vecset3dshaperepresentation}, CLAY~\cite{zhang2024claycontrollablelargescalegenerative}, and TripoSG~\cite{li2025triposghighfidelity3dshape} and shares the same architecture with CLAY's VAE. To transform 3D geometry into latent space, we first sample a point cloud $X$ from 3D geometry's surface and adopt a two-stage downsampling process (random downsample firstly then apply farthest point sampling~\cite{garland1997surface}) for $X$ to get query points $X_0$. Lastly, the cross attention is applied to $X$ and $X_0$ with learnable positional embeddings to obtain the latent code $z$.:
{\small \begin{equation}
    z = \mathcal{E}(X) = \text{CrossAttn}(\text{PosEmb}(X_0), \text{PosEmb}(X))
\end{equation}}

The 3D-VAE decoder, composed of several self attention layers and a cross attention layer, transforms the latent code $z$ to a signed distance field (SDF) with preset query points $q$.:
{\small \begin{equation}
    SDF = \mathcal{D}(q, z) = \text{CrossAttn}(\text{PosEmb}(q), \text{SelfAttn}(z))
\end{equation}}

\noindent\textbf{Diffusion Transformer.} Our DiT network, comprised of 24 DiT blocks with a width of 768 and totaling 368M parameters, is designed with efficiency in mind. Each DiT block consists of a multi-head cross-attention layer (MCA), a multi-head self-attention layer (MSA), and a feedforward network (FFN), interleaved with layer normalization (LN). To accommodate the relatively small width, we inject the time step $t$ using the adaptive layer normalization (adaLN)~\cite{perez2017filmvisualreasoninggeneral}, modulating MSA, MCA, and FFN via
factors \( g \), \( \gamma \), and \( s \), obtained by a MLP conditioned on $t$.:
\vspace{-3pt}
{\small \begin{equation}
    z = z + g_{\text{msa}} \cdot \text{SelfAttn}(\text{mod}(\text{LN}_{\text{msa}}(z), s_{\text{msa}}, \gamma_{\text{msa}}))
\end{equation}}

\noindent Turning to conditioning on the image prompt $y$, we adopt cross-attention to ensure spatial alignment between latent code and image feature, which is extracted by DINOv2~\cite{oquab2024dinov2learningrobustvisual}. The conditioning is defined as:
\vspace{-3pt}
{\small \begin{equation}
    z = z + g_{\text{mca}} \cdot \text{CrossAttn}(\text{mod}(\text{LN}_{\text{mca}}(z), s_{\text{mca}}, \gamma_{\text{mca}}), y)
\end{equation}}

\noindent At last, tokens will pass through a feedforward network:
\vspace{-3pt}
{\small \begin{equation}
    z = z + g_{\text{ffn}} \cdot \text{FFN}(\text{mod}(\text{LN}_{\text{ffn}}(z), s_{\text{ffn}}, \gamma_{\text{ffn}}))
\end{equation}}

\noindent\textbf{Rectified Flow and Loss Functions.} Unlike previous generative models modeling the mapping between a Gaussian distribution and the ground truth data distribution, we model the mapping between the coarse geometry distribution and the fine geometry distribution. Let \( z_1 \) represent the fine geometry's latent code and \( z_0 \) corresponds to the coarse geometry's latent code. The geometry latent code at time \( t \), \( z_t \), is given by:
\vspace{-2pt}
{\small \begin{equation} 
z_t = (1 - t) \cdot z_0 + t \cdot z_1
\end{equation}}
\vspace{-2pt}
The conditional probability \( P(z_1 | z_t) \) is defined as:
\vspace{-2pt}
{\small \begin{equation}
P(z_1 | z_t) = \mathcal{N}(z | (1 - t) \cdot z_0 + t \cdot z_1, \sigma_{\min}^2 \cdot I)
\end{equation}}
\vspace{-2pt}
The loss function \( L(\theta) \) is:
\vspace{-2pt}
{\small \begin{equation}
L(\theta) = \mathbb{E}_{t, z_0, z_1, y} \left[ \left\| v_\theta(t, z_t, y) - (z_1 - z_0) \right\|^2 \right]
\end{equation}}
\vspace{-2pt}
where \( v_\theta \) is parameterized by the DiT with learnable parameters \(\theta\) and \(y\) is the image prompt. The L2 loss constrains the prediction of DiT's prediction $v_\theta$ with respect to the ground truth velocity $(z_1 - z_0)$ in latent space.

\noindent\textbf{Noise Augmentation.} Since the degraded geometric distribution is relatively independent, direct modeling of distribution transformation between coarse and fine geometry cannot effectively learn comprehensive refinement rules from the dataset, leading to noisy results. Inspired by~\cite{fischer2024boostinglatentdiffusionflow}, we apply noise augmentation to the latent code of coarse geometry, which helps improve the quality. The noise \( \epsilon \) is added to \( z_0 \) before obtaining \( z_t \) instead of after, as applying noise after obtaining \( z_t \) would affect the efficiency of optimal transport, leading to an optimization path that is not straight. See supplementary material for more details.

\subsection{Token Matching}
We propose token matching, a training method to match the latent codes of coarse and fine geometries for efficient training, as shown in Fig~\ref{fig:inference_pipeline} (3). Training without token matching leads the network to learn not only refinement rules but also unnecessary operations (e.g., swapping latent codes), significantly reducing training efficiency. However, measuring the similarity between latent codes is non-trivial. Since the latent code represents the space area around the query points $X_0$ (Fig~\ref{fig:inference_pipeline} (4)), we are inspired to match latent codes in latent space similarly to matching query points in 3D space.

To balance effectiveness and efficiency, we propose a novel token matching method. First, we obtain the query points of fine geometry following 3D-VAE's two-stage downsampling (applying random downsampling then farthest point sampling on fine geometry's point cloud). 
Next, we obtain the coarse geometry's query points by applying a nearest-neighbour algorithm. Specifically, for each fine query point $X_0^\text{fine}$, we identify its closest point within the full coarse point cloud. These identified points from the coarse cloud serve as the coarse query points $X_0^\text{coarse}$, establishing a spatial correspondence with $X_0^\text{fine}$. 
This method is computationally efficient, maintaining model generalization and offering robust performance across both simple and complex geometries while ensuring local refinement. We further discuss token matching in the supplementary material.

\subsection{Data Curation}


To achieve effective local transformation between coarse and fine geometries, well‐aligned coarse–fine pairs are crucial. Traditional geometric degradation techniques—such as Taubin smoothing~\cite{466848}- can effectively reduce high‐frequency noise in complex geometries but tend to be too subtle for simpler ones. Moreover, applying the same degradation across all objects risks distorting complex shapes while failing to degrade simpler ones sufficiently. To address these issues, we adopt the reconstruction outputs from an LRM as shown in Fig~\ref{fig:inference_pipeline} (2), which naturally introduce a balanced level of geometric degradation that adapts to an object’s inherent complexity. This choice not only preserves spatial correspondence but also better replicates the artifacts typical of neural reconstructions. Further details of data curation strategy are provided in the supplementary material.

Considering that fine objects are essential for training, we select Objaverse as our dataset and filter out the low-quality objects by evaluating each object's quality based on CLIP~\cite{radford2021learningtransferablevisualmodels} feature of four ortho rendering views. After filtering, our training set consists of about 110,000 objects, most of them are rich in details. We randomly partition the dataset into a training set and an evaluation set, with the latter containing 350 objects. The coarse geometry is reconstructed using a reimplemented version of Instant3D~\cite{li2023instant3dfasttextto3dsparseview}, leveraging four orthographic views of objects at a resolution of 512.

\vspace{-0.5em}
\section{Experiment}
\label{sec:Experiment}

\setlength{\parindent}{2em}



\begin{figure*}[htbp]
\vspace{-2em}
    \centering
    \includegraphics[width=\textwidth]{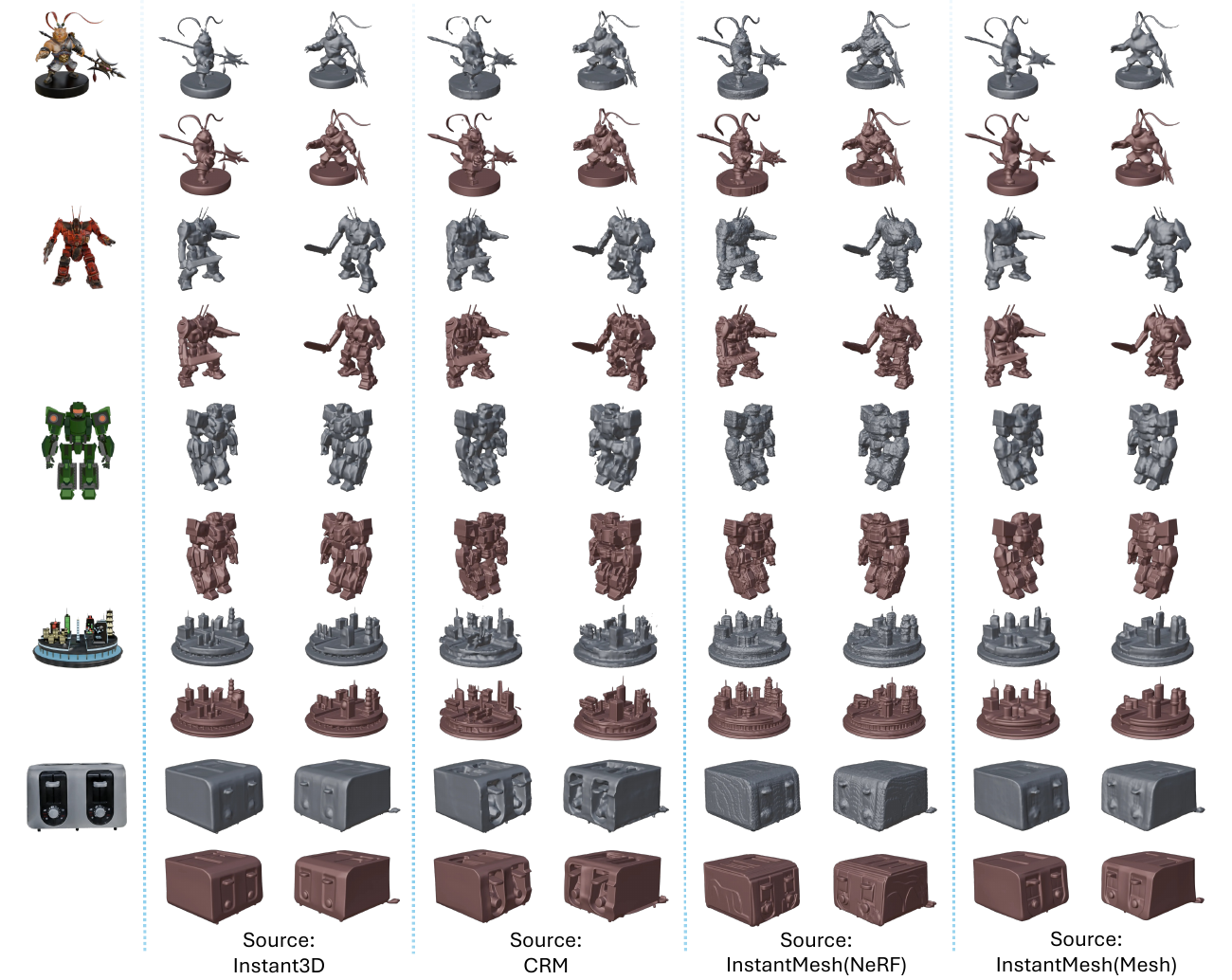} 
    \caption{We apply our method on input meshes reconstructed by different approaches (Instant3D \cite{li2023instant3dfasttextto3dsparseview}, CRM \cite{wang2024crmsingleimage3d}, InstantMesh \cite{xu2024instantmeshefficient3dmesh}). \textcolor{coarse_color}{\rule{0.8em}{0.8em}} represent coarse, \textcolor{fine_color}{\rule{0.8em}{0.8em}} represent fine refinement results from our method. The top three objects are from Objaverse, while the bottom object is from GSO. More results can be found in the supplementary.}
    \label{fig:recon}
\end{figure*}
\begin{figure*}[htbp]
\vspace{-2em}
    \centering
    \includegraphics[width=\textwidth]{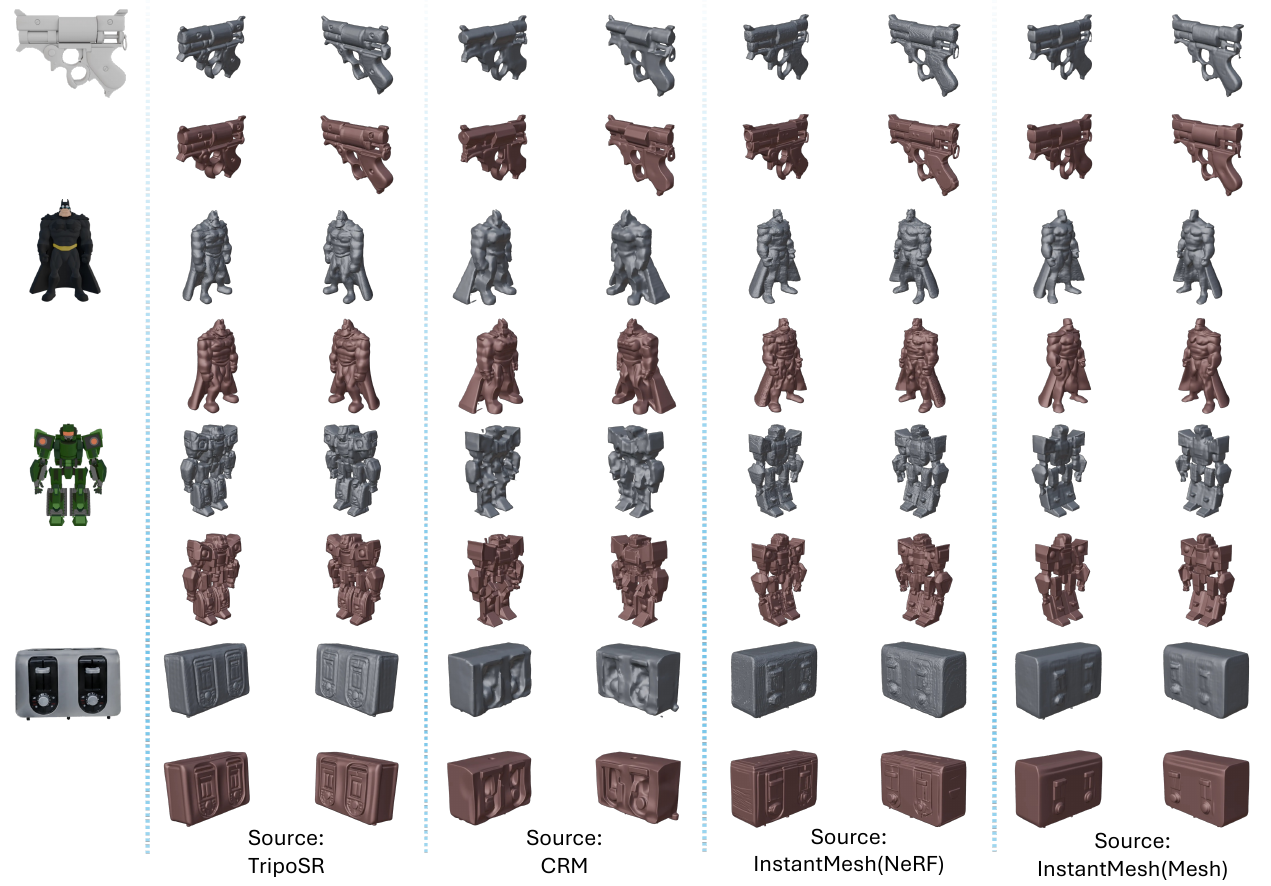} 
    \caption{We apply our method on input meshes generated by different approaches (TripoSR~\cite{tochilkin2024triposrfast3dobject}, CRM~\cite{wang2024crmsingleimage3d}, InstantMesh~\cite{xu2024instantmeshefficient3dmesh}). \textcolor{coarse_color}{\rule{0.8em}{0.8em}} represent coarse, \textcolor{fine_color}{\rule{0.8em}{0.8em}} represent fine refinement results from our method. The top three objects are from Objaverse, while the bottom object is from GSO. More results can be found in the supplementary.}
    \label{fig:generate}

\end{figure*}

Our comprehensive evaluation of the model's 3D geometric refinement capabilities is divided into three parts:
1). \textit{FeedForward-based Reconstruction}:
We tested the refinement ability of our method on the reconstruction results of various LRM models to demonstrate its effectiveness.
2). \textit{Generation:}
We assessed our method's performance on the generation outputs from different LRM models, showcasing its versatility across diverse 3D representations. 3). \textit{More 3D representations:}
We applied our method to reconstruction results reconstructed by NeuS-like~\cite{wang2023neuslearningneuralimplicit} optimization-based and generation results generated by Rodin Gen-1\footnote{\url{https://hyperhuman.deemos.com/rodin}} and Neural4D\footnote{\url{https://www.neural4d.com/}}, achieving impressive results that highlight its strong generalization. We also performed ablation studies on our design choices. See the supplementary for more implementation and experiment details.

For the metric, we select FID~\cite{heusel2017gans} as our metric and obtain each object's FID result following the setting of SDF-styleGAN~\cite{zheng2022sdfstylegan}, which can assess visual quality, capture statistical differences, detect mode collapse, and serve as a standardized benchmark despite not measuring fine-grained consistency perfectly. Considering FID may not fully capture the perceptible difference in fine details, we also mixed results from different datasets, methods, and tasks for a user study. 




\subsection{Feed-forward Reconstruction}
\label{sec:Feed-forward Reconstruction}


Considering the coarse-fine pairs used in the training process, where all coarse models are reconstructed by Instant3D~\cite{li2023instant3dfasttextto3dsparseview}, this experiment aims to evaluate the generalizability to refining the reconstruction results of different kinds of LRM, including Instant3D as well.

For the evaluation set, we randomly sampled 199 models from the GSO~\cite{downs2022googlescannedobjectshighquality} dataset and selected 350 unseen models from~\cite{deitke2022objaverseuniverseannotated3d} as the evaluation set. For the LRM, We used Instant3D~\cite{li2023instant3dfasttextto3dsparseview}, CRM~\cite{wang2024crmsingleimage3d}, InstantMesh~\cite{xu2024instantmeshefficient3dmesh} (testing both NeRF and Mesh, the checkpoints used are all the official large versions provided). We render multi-view images according to different LRM's requirements and use their reconstruction results as coarse model.


Visualization results can be found in Fig~\ref{fig:recon} and FID results can be found in Tab~\ref{tab:FID_result}. Our method can highlight the blurry details in the coarse model and add details only exist in the image prompt. In addition, for the aliasing existing on the input geometric surface, our method can eliminate them and show a smoother geometric surface. The FID measurement results reflect the effectiveness and robustness of our method for different models and different objects. 


\begin{table}[htbp]
\vspace{-1em}
    \centering
    \resizebox{\linewidth}{!}{%
        \begin{tabular}{c|cc|cc}
            \toprule
            \multirow{2}{*}{\centering FID $\downarrow$} & \multicolumn{2}{c|}{Reconstruction} & \multicolumn{2}{c}{Generation} \\
            & Coarse & Fine & Coarse & Fine \\

            \midrule
            Instant3D~\cite{li2023instant3dfasttextto3dsparseview} & 20.33 & \textbf{19.07} & \textsf{x} & \textsf{x} \\
            TripoSR~\cite{tochilkin2024triposrfast3dobject} & \textsf{x} & \textsf{x} & 51.80 & \textbf{33.33} \\

            CRM~\cite{wang2024crmsingleimage3d} & 40.13 & \textbf{25.29} & 48.91 & \textbf{33.74} \\

            InstantMesh (NeRF)~\cite{xu2024instantmeshefficient3dmesh} & 58.35 & \textbf{29.51} & 50.12 & \textbf{25.75} \\
            InstantMesh (Mesh)~\cite{xu2024instantmeshefficient3dmesh} & 35.32 & \textbf{24.45} & 33.85 & \textbf{25.79} \\

            \bottomrule
        \end{tabular}%
    }
    \caption{FID comparison of applying our method to different LRM results on reconstruction/generation task using Objaverse.}
    \label{tab:FID_result}
    \vspace{-1em}
\end{table}

\begin{figure*}[htbp]
    \centering
    \includegraphics[width=\textwidth]{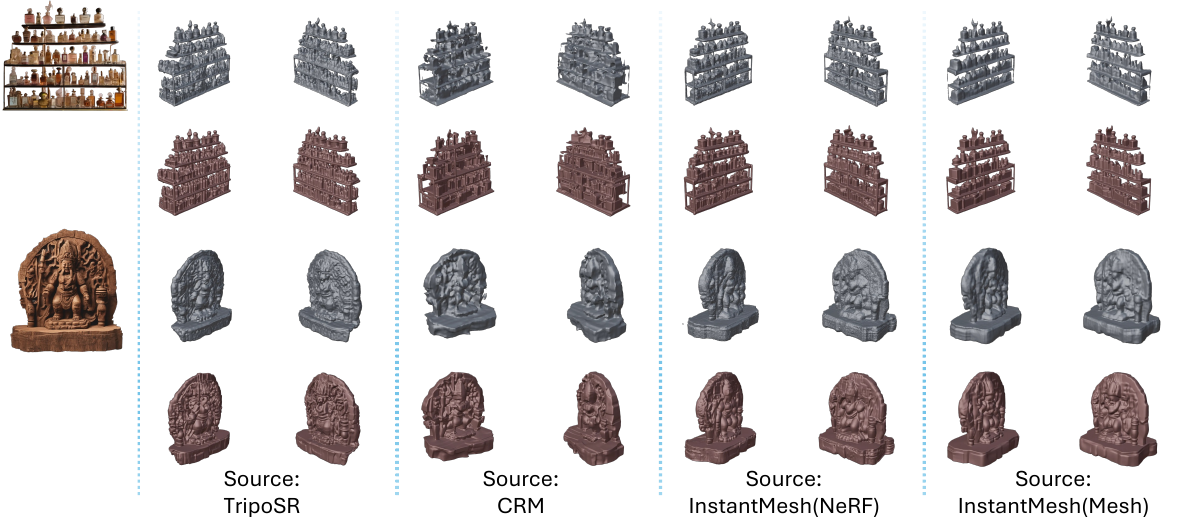} 
    \caption{We apply our method on input meshes generated by different approaches (TripoSR~\cite{tochilkin2024triposrfast3dobject}, CRM~\cite{wang2024crmsingleimage3d}, InstantMesh~\cite{xu2024instantmeshefficient3dmesh}) using GPTEval3D as input.  \textcolor{coarse_color}{\rule{0.8em}{0.8em}} represent coarse, \textcolor{fine_color}{\rule{0.8em}{0.8em}} represent fine refinement results from our method. More results can be found in the supplementary.}
    \label{fig:paper_generate_gpteval3d}
\end{figure*}

\vspace{-1em}
\subsection{Generation}

For the generation refinement task, we randomly sampled 199 models from the GSO~\cite{downs2022googlescannedobjectshighquality} dataset and selected 350 unseen models from Objaverse~\cite{deitke2022objaverseuniverseannotated3d} as the evaluation set, same as mentioned before. 
Additionally, to explore the refinement ability, we used 110 images provided by GPTEval3D~\cite{wu2023gpteval3d} as input, employing TripoSR~\cite{tochilkin2024triposrfast3dobject}, CRM~\cite{wang2024crmsingleimage3d} and InstantMesh~\cite{xu2024instantmeshefficient3dmesh} (testing both NeRF and Mesh, the checkpoints used are all the official large versions provided.) as generation models for the coarse geometry. 

As the visualization results illustrated in Fig~\ref{fig:generate} and FID results shown in Tab~\ref{tab:FID_result}, our method performs particularly well on complex objects, while coarse models' surface have large geometric noise and lack details. Furthermore, our method shows outstanding performance in FID scores.

For generation results, some of them are not well aligned with the input image. However, our method can still refine them in a proper way, where aligning the original shape and adding finer local detail. It shows that although our method learns refinement ability with aligned image-coarse shape pairs, our method knows how to use image prompt to add local details.

For experiments on GPTEval3D generation results, refinement results shows the robustness of our method to such challenging data, as illustrated in Fig~\ref{fig:paper_generate_gpteval3d}. Each image from GPTEval3D~\cite{wu2023gpteval3d} is used to obtain coarse models through image-to-3D models (i.e., TripoSR, CRM, InstantMesh) and serves as an image prompt in the refinement process. Due to the high level of difficulty, most image-to-3D generation models struggled to produce reasonable 3D models based on these images. We selected the plausible 3D models from them and applied our refinement method.


\subsection{More 3D representations}


\begin{figure*}[h]
    \centering
    \includegraphics[width=\textwidth]{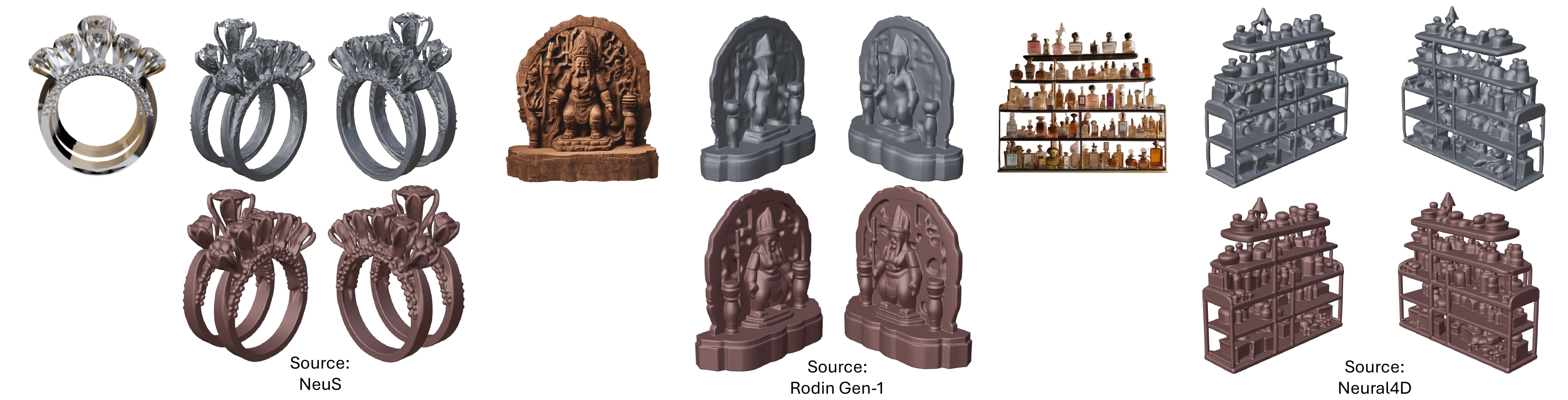} 
    \caption{We apply our method on input meshes reconstructed by NeuS and generated by Rodin Gen-1 and Neural4D. \textcolor{coarse_color}{\rule{0.8em}{0.8em}} represent coarse, \textcolor{fine_color}{\rule{0.8em}{0.8em}} represent fine refinement results from our method. More results can be found in the supplementary.}
    \label{fig:neus_Rodin_Neural4D}
\end{figure*}

\begin{table*}[t]
    \centering
    \resizebox{\textwidth}{!}{  
        \begin{tabular}{c|c|c|c|c|c|c|c|c}
            \toprule
            \textbf{Method} & \textbf{Coarse} & \textbf{w/o Token matching} & \textbf{w/o Image condition} & \textbf{w/o Noise augmentation} & \textbf{Cross attention} & \textbf{Sorting} & \textbf{w/o training} & \textbf{Our full method} \\
            \midrule
            \textbf{FID $\downarrow$} & 20.33 & 19.79 & 21.31 & 17.98 & 23.92 & 20.58 & 24.22 & 19.32 \\
            \bottomrule
        \end{tabular}
    }
    \caption{FID scores evaluated on the Objaverse evaluation set (350 objects) for different model ablations.}
    \label{tab:ablation_study}
\end{table*}

\begin{figure}
    \centering
    \captionsetup{type=figure}
    \includegraphics[width=\linewidth]{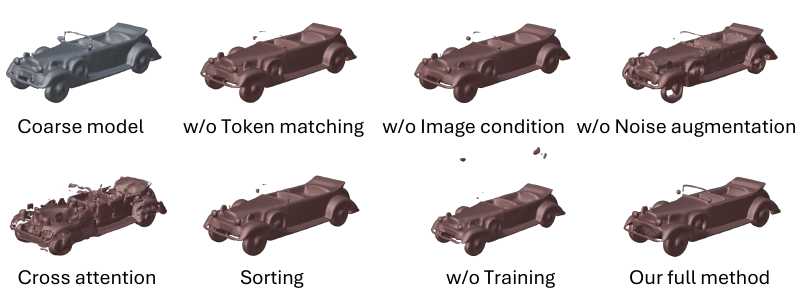} 
    \captionof{figure}{Visualization of inference results with different settings.}
    \label{fig:ablation_study}
\end{figure}

To further evaluate the generalizability to other 3D representations, we tested our refinement ability on reconstruction results of NeuS~\cite{wang2023neuslearningneuralimplicit} and generation results of two commercial products, i.e., Rodin Gen-1, Neural4D.

Considering the long optimization time of NeuS and the limited quota of Rodin Gen-1 and Neural4D, randomly sampled a subset of the evaluation set of Objaverse and GSO used in previous experiments for evaluation.


The refinement results are all illustrated in Fig~\ref{fig:neus_Rodin_Neural4D} and demonstrate that our method successfully refines and corrects irregular details. The results of NeuS are reconstructed with 40 views evenly distributed around each object. However, due to the lack of prior guidance, they are prone to artifacts in scenarios with complex occlusions, and the smoothness of the surface remains imperfect. After refining by our method, the shredded parts become continuous, and details are more obvious. Since commercial products are typically trained on huge 3D data with a large number of GPUs, we selected images from GPTEval3D as input. The generation results have smooth surface but still lack details. After refinement, more details are introduced, and the shapes become more plausible.


\subsection{User Study}

The FID metric, while showing incremental improvement, may not fully capture the perceptible difference in fine details achieved by our method, especially on complex geometries. We invited 32 researchers, primarily experts in 3D vision, and compared the geometry quality of 10 models (reconstructed and generated by Instant3D~\cite{li2023instant3dfasttextto3dsparseview}, InstantMesh~\cite{xu2024instantmeshefficient3dmesh}, CRM~\cite{wang2024crmsingleimage3d}, TripoSR~\cite{tochilkin2024triposrfast3dobject}) before and after refinement. In the questionnaire, we presented four views (azimuths: 45°, 135°, 225°, 315°; elevation: 15°) of both coarse and refined geometries, rendered with consistent code to ensure varied order and same color to avoid bias, and asked users to identify which of the two models (coarse vs. refined) appeared more detailed and closer to the input image. Statistics show that in 94\% of the comparisons, participants judged the geometry obtained by our refinement method to be better (more detailed and closer to the image). This strongly validates the perceptual effectiveness of our method's refinement ability.

\subsection{Ablation Study}

In our ablation study, we tested the effects of removing token matching, using cross-attention to inject coarse geometry information, removing image prompts, removing noise augmentation, and using a point cloud sorting algorithm to query point cloud, inference without training for reference. For all ablation experiments, we trained for approximately 100,000 steps on a dataset of 40,000 objects (subsample from the training set), with the results tested on an evaluation set of 350 objects from Objaverse (same as in previous experiments). We used FID as the evaluation metric on the rendered images following the SDF-StyleGAN~\cite{zheng2022sdfstylegan} settings. The FID results are shown in Tab~\ref{tab:ablation_study}, and the visualizations are presented in Fig~\ref{fig:ablation_study}.


Experimental results indicate that removing token matching slows convergence, as the model must simultaneously learn both refinement transformations and unnecessary operation (e.g., position swapping); it is hard to generate unseen details in the coarse geometry without the image condition; cross-attention modeling lacks the constraint of one-to-one token correspondence between coarse and fine models, making the noise-to-fine process inefficient and yielding worse geometric quality after refinement; applying point cloud sorting to the query point cloud yields no clear benefit, and inference without training is almost as effective in predicting coarse geometry.

Removing noise augmentation results in the lowest FID, as shown in Tab~\ref{tab:ablation_study}. We speculate that this is primarily due to two reasons: first, the Inception model~\cite{szegedy2016rethinking}, which computes FID, is not sensitive to the noise perception of the ensemble model, and second, the image conditioning ensures the results are more consistent with the ground truth, even without noise augmentation. While these factors contribute to low FID, the visual degradation remains clearly noticeable, as demonstrated in Fig~\ref{fig:ablation_study} and further illustrated in the supplementary material.


\vspace{-0.5em}

\setlength{\parindent}{2em}

\section{Discussion and Conclusion}
\label{sec:Conclusion}



\noindent\textbf{Conclusion.} In this paper, we present a generative 3D geometry refinement method using an image as a prompt. We introduce a training technique called Token Matching for localized geometry refinement. This approach proves highly effective in both reconstruction and generation tasks, delivering refined results across diverse datasets, particularly for complex geometries.



\clearpage
\setcounter{page}{1}
\maketitlesupplementary


    

\begin{figure*}[ht] 
\centering
\begin{minipage}{\textwidth} 
\begin{algorithm}[H]
\caption{Data-Dependent Rectified Flow}
\begin{algorithmic}[1]
\Procedure{$\mathcal{Z}$}{$\text{RectFlow}((X_0, X_1))$}

    \State \textit{Inputs}: Draws from a coupling $(X_0, X_1)$ of $\pi_0$ and $\pi_1$; velocity model $v_\theta: \mathbb{R}^d \to \mathbb{R}^d$ with parameter $\theta$.

    \State \textit{Training:}
    \State 1. Obtain $\pi_0$ and $\pi_1$ using $\pi_0 = \mathcal{E}(G_0)$ and $\pi_1 = \mathcal{E}(G_1)$, where $\mathcal{E}$ denotes the VAE encoder, $G_0$ represents the coarse geometry, and $G_1$ represents the fine geometry.
    \State 2. Optimize $\hat{\theta} = \arg\min_\theta \mathbb{E} \big[ \| X_1 - X_0 - v(tX_1 + (1-t)X_0, t) \|^2 \big]$, with $t \sim \text{Uniform}([0, 1])$.

    \State \textit{Sampling:}
    \State 1. Start with $Z_0 \sim \pi_0$, where $\pi_0 = \mathcal{E}(G_0)$. Here, $\mathcal{E}$ denotes the VAE encoder, $G_0$ represents the coarse geometry.
    \State 2. Generate $(Z_0, Z_1)$ by solving $\mathrm{d}Z_t = v_{\hat{\theta}}(Z_t, t) \mathrm{d}t$, obtaining $\{Z_t : t \in [0, 1]\}$.

    \State \textit{Return}: $\mathcal{Z} = \{Z_t : t \in [0, 1]\}$.
\EndProcedure
\end{algorithmic}
\end{algorithm}
\end{minipage}
\end{figure*}


In this supplement, we first provide the implementation detail in Sec.~\ref{sec:Implementation Details}. We also provide further
experiment details in Sec.~\ref{sec:Experiment Details} and further discussion in Sec.~\ref{sec:Further Discussion}. Finally, we provide additional visual results in Sec.~\ref{sec:Additional Visual Results}. We encourage the readers to view our accompanying videos in the supplement, showcase the rotation of objects rendered with normals as presented in the paper.

\section{Implementation Details}
\label{sec:Implementation Details}





\subsection{Model architecture}
For the 3D-VAE encoder, after sampling $N$ points from geometry's surface, we first randomly downsample it to $4 \times M$, where $M$ is the number of latent codes of each object and $M = 2048$, $N = 20480$. Next, obtaining query points $X_0$ by using farthest point sampling to further subsample it with $\frac{1}{4}$ ratio. 

The 3D-VAE decoder, comprised 24 multi-head self attention layer and a multi-head cross attention layer. For the preset query points used in cross attention layer, it evenly distributed in 3D space, using for querying the corresponding spatial SDF value, which can adopt marching cube algorithm to convert it into mesh format.

For the Refinement DiT is comprised 24 DiT blocks with a width of 768, 12 attention heads, and a latent length of 2048, totaling 368M parameters. For the image prompt, we use the image feature extracted by DINOv2~\cite{oquab2024dinov2learningrobustvisual}. The feedforward network in each DiT block consists of a two layer multi layer perception with GELU activation and the middle dimension is four times the input dimension.

\subsection{Noise Augmentation} During training, we apply noise to \( z_0 \) according to the DDPM~\cite{ho2020denoisingdiffusionprobabilisticmodels} linear schedule at 400 timesteps. During the inference stage, noise augmentation is optionally, which can reduce the impact of floating objects in the coarse model on the final refined results. For our experiment, we all add noise at 100 time steps.

\subsection{Data Curation} To construct satisfying coarse-fine pairs, we choose to obtain coarse model by using Instant3D to reconstruct through four ortho views at a resolution of 512. It is worth mention that some of the objects are textureless. For the training dataset, we select Instant3D~\cite{li2023instant3dfasttextto3dsparseview} to obtain coarse model, and the objects are all from Objaverse~\cite{deitke2022objaverseuniverseannotated3d}. It is worth noting that the Instant3D we use is reimplemented by us because its code has not been released yet.

In order to align the coarse geometry with the fine geometry in space, we first normalize and rescale each object to fit within a bounding box of side length 1, then translate the object so that the bounding box is centered at the origin. 


\subsection{Training}

In our training setup, we start with a learning rate of 1e-10 and warm it up to 1e-4 over 5,000 steps. We use a total batch size of 256. We train our DetailGen3D model for 1,000 epochs, which takes approximately eight days on eight A800 GPUs. When training the data-dependent rectified flow, we randomly zero the DINOv2 features with a probability of 10\% to enable classifier-free guidance during inference, thereby improving the quality of conditional generation. For the DINOv2~\cite{oquab2024dinov2learningrobustvisual} checkpoint, we use the ViT-L/14 distilled with the registered version, downloaded from the official DINOv2 GitHub repository\footnote{\url{https://github.com/facebookresearch/dinov2}}. For the image prompt, we select a forward view with same camera intrinsic and extrinsic for training because using more views or highly random camera poses leads to longer training time. We believe that introducing camera pose embedding (e.g., plucker embedding) will help.




\section{Experiment Details}
\label{sec:Experiment Details}


\begin{figure}[htbp]
    \centering
    \includegraphics[page=1,width=\columnwidth]{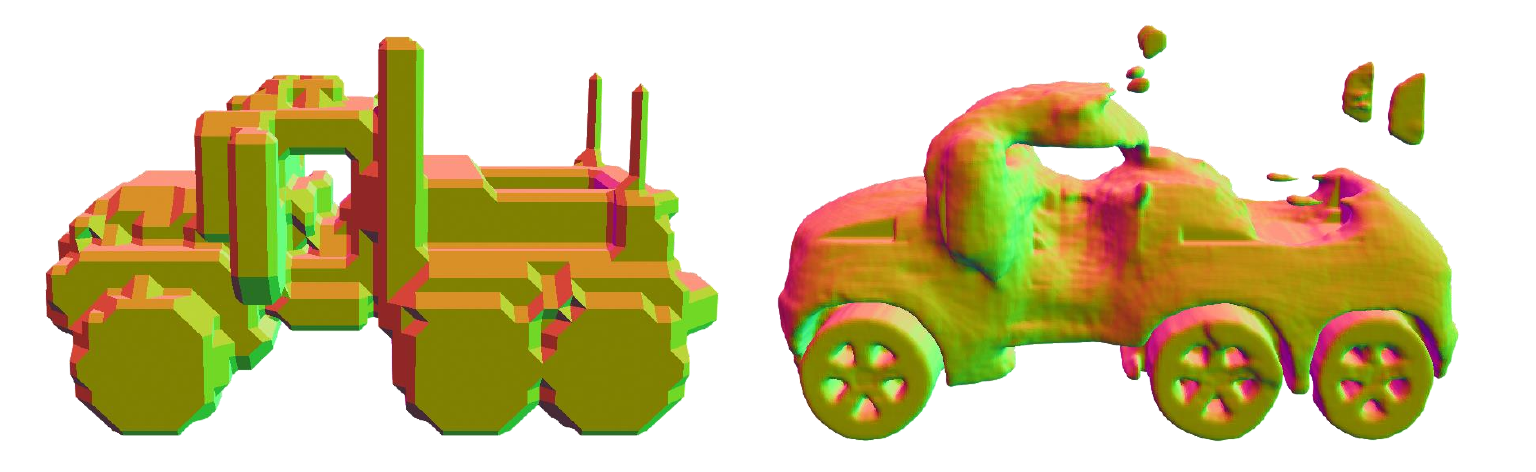} 
    \caption{Coarse model (left) is degraded and rotated manually following ShaDDR training setting. The model refined by ShaDDR (right) is worse.}
    \label{fig:ShaDDR}
\end{figure}

\subsection{Other Methods}
We didn't compare with other shape detailization methods as listed in related work because our tasks are different and they cannot handle our evaluation set, as illustrated in Fig.A~\ref{fig:ShaDDR}. They aim at increasing the resolution of extremely low resolution voxels to relative higher resolution (e.g., $16^3$ to $64^3$). Our evaluation set consists of up to 450 objects, which are comes from GSO and Objaverse, however, their evaluation set only consists of tens objects and comes from the same category in ShapeNet~\cite{chang2015shapenetinformationrich3dmodel} and have strict requirements on the orientation of objects (Refining a car from Objaverse using ShaDDR~\cite{chen2023shaddrinteractiveexamplebasedgeometry}'s checkpoint—which is trained on car objects from ShapeNet—produces worse results due to the difference in orientation compared to the training set.) and the input's orientation shown in~\ref{fig:ShaDDR} is manually adjusted to meet ShaDDR's requirements. 

\subsection{Evaluation Dataset}
For the Objaverse~\cite{deitke2022objaverseuniverseannotated3d} evaluation set used, the IDs for models trained with different LRM methods are varied and not publicly available, which may lead to unfair comparisons between methods. However, we emphasize that our refinement model has never seen these 3D models during training, so its ability to refine the same object across different models fairly reflects the model's generalization capability.



\subsection{SDF-stylegan Setting}
For the feed-forward reconstruction and generation experiment, the FID~\cite{heusel2017gans} metric we evaluate following SDF-stylegan~\cite{zheng2022sdfstylegan}, rendering 20 views with preset random camera poses and same color (grey).

\subsection{NeuS Reconstruction}
 For NeuS~\cite{wang2023neuslearningneuralimplicit}, to further improve speed, we used Instant-NSR~\cite{instant-nsr-pl} implementation. For the multi-view data, we rendered a uniformly distributed set of 40 views as input. The camera poses are elevation with -60°, -30°, 0°, 30°, 60° and azimuth with 0°, 45°, 90°, 135°, 180°, 225°, 270°, 315°. 

\section{Further Discussion}
\label{sec:Further Discussion}
\subsection{Token Matching}
Although attempts were made to match latent code, the correspondence is difficult to model effectively. We explored the use of point cloud sorting algorithms, which work well for simple geometries but struggle with complex geometries due to their inability to preserve spatial relationships. Optimal Transport (OT) between coarse and fine query points can handle more complex shapes, but is computationally expensive and must be performed offline.

It is worth mentioning that using a learnable query in place of query points in the VAE does not solve the problem. The first reason is that the VAE quality with learnable queries is lower than that with query points. The second reason is that using learnable queries makes it impossible to ensure that the tokens with the same indices in the tensor, obtained after encoding the coarse geometry and fine geometry, have similar meanings.

Finally, we match the coarse geometry's latent code with fine geometry's latent code by applying nearest-neighbour algorithm to identify the closest fine geometry query points in the coarse geometry point cloud, using these as the coarse geometry's query points. While this design cannot theoretically guarantee one-to-one correspondence, it does so experimentally. This is because the 3D-VAE's two-stage downsampling process (i.e., downsampling point cloud $X$ sampled from geometry surface to query points $X_0$)—random downsampling followed by farthest point sampling—results in query points $X_0$ of fine geometry being distributed far apart in space. As a result, the nearest-neighbour algorithm becomes a viable method for matching latent codes.

\subsection{Data Curation}
While our experimental results have shown our method has strong generalizability across different sources of coarse models, including both generation and reconstruction tasks, there is still room for improvement. Using only one type of LRM (e.g., Instant3D) may introduce bias, and we believe that applying geometry degradation and mixing multiple LRM reconstruction results will be of help. 

\subsection{Application}
Our focus on geometry refinement stems from the fact that in many applications (e.g., design, simulation) replying on fine geometry, whereas color can be integrated later~\cite{ren2024xcubelargescale3dgenerative, wu2024direct3d}. For the texture, our method supports to simply reproject the original textures onto the refined mesh or use separate texture-generation pipelines~\cite{zeng2023paint3dpaint3dlightingless,Yu_2024}. This ensures high-quality geometry while preserving the flexibility to include color information as needed.

\subsection{Limitation.} Our method can robustly refine geometry with image prompt. However, it still struggles with ultra-precision details like extremly complex and small geometry shown in image. This issue might be solved by training a better 3D-VAE, collecting more diverse coarse and fine geometry pairs for training.

\section{Additional Visual Results}
\label{sec:Additional Visual Results}
Due to the extensive nature of the content, the complete details and supplementary materials are available in the full appendix on our homepage. For further information, please visit:\href{https://detailgen3d.github.io/DetailGen3D/}{DetailGen3D}
\newpage

{
    \small
    \bibliographystyle{ieeenat_fullname}
    \bibliography{main}
}


\end{document}